\documentclass[11pt]{article}

\usepackage[final]{acl}

\usepackage{times}
\usepackage{bm}
\usepackage{amsmath}
\usepackage{latexsym}
\usepackage{booktabs}
\usepackage{multirow}
\usepackage{pifont}
\usepackage{colortbl}
\usepackage{arydshln}
\usepackage{makecell}
\usepackage{fancyvrb}
\usepackage{siunitx}
\usepackage{float}

\newcounter{paranum}
\newcommand{\paranum}{\refstepcounter{paranum}}
\newcommand{\labelpara}[1]{\paranum\label{#1}}

\usepackage[T1]{fontenc}

\usepackage[utf8]{inputenc}

\usepackage{microtype}

\usepackage{inconsolata}

\usepackage{graphicx}

\title{Comprehensive and Efficient Distillation for Lightweight Sentiment Analysis Models}

\author{
Guangyu Xie$^{1}$\thanks{\quad The first two authors contribute equally to this work.}\ , Yice Zhang$^{1\ast}$, Jianzhu Bao$^{3,1}$,\\ 
\bf Qianlong Wang$^{1}$, Yang Sun$^{1}$, Bingbing Wang$^{1}$, and Ruifeng Xu$^{1,2}$\thanks{\quad Corresponding Authors}\\
 $^{1}$ Harbin Institute of Technology, Shenzhen, China \\
 $^{2}$ Peng Cheng Laboratory, Shenzhen, China \\
 $^{3}$ Nanyang Technological University, Singapore \\
\texttt{guangyuxie2001@gmail.com,zhangyc\_hit@163.com,xuruifeng@hit.edu.cn} \\
}

\begin{document}
\maketitle
\begin{abstract}

Recent efforts leverage knowledge distillation techniques to develop lightweight and practical sentiment analysis models. These methods are grounded in human-written instructions and large-scale user texts.  
Despite the promising results, two key challenges remain: (1) manually written instructions are limited in diversity and quantity, making them insufficient to ensure comprehensive coverage of distilled knowledge; (2) large-scale user texts incur high computational cost, hindering the practicality of these methods.
To this end, we introduce \textsc{CompEffDist}, a comprehensive and efficient distillation framework for sentiment analysis. Our framework consists of two key modules: attribute-based automatic instruction construction and difficulty-based data filtering, which correspondingly tackle the aforementioned challenges. Applying our method across multiple model series (Llama-3, Qwen-3, and Gemma-3), we enable 3B student models to match the performance of 20x larger teacher models on most tasks.
In addition, our approach greatly outperforms baseline methods in data efficiency, attaining the same performance level with only 10\% of the data. All codes are available at \url{https://github.com/HITSZ-HLT/COMPEFFDIST}.

\end{abstract}

\section{Introduction}

Recent research shows that large language models (LLMs) possess robust sentiment analysis capabilities.
Without requiring task-specific fine-tuning, these models can achieve exceptional performance across various sentiment-related tasks, including polarity determination \cite{zhang-etal-2024-sentiment}, emotion recognition \cite{10.1145/3637528.3671552}, sarcasm detection \cite{Yao_Zhang_Li_Qin_2025}, and stance detection \cite{zhang2024stancedetectiontechniquesevolve}. Furthermore, these models demonstrate strong abilities to reason and interpret sentiments in complex contexts \cite{fei-etal-2023-reasoning,zhang2024distillingfinegrainedsentimentunderstanding}. 

Despite remarkable performance, the substantial parameter size of LLMs severely constrains their practical application. To address this limitation, extensive research \cite{zhong-etal-2024-revisiting,gu2024minillm,10.5555/3692070.3693067,wu-etal-2024-lamini,peng2024pretrainingdistillationlargelanguage} has focused on leveraging knowledge distillation techniques \cite{hinton2015distillingknowledgeneuralnetwork} to transfer knowledge and skills from large teacher models to more compact student models, thereby reducing deployment costs. Among these studies, targeted distillation \cite{liu2023tinygsmachieving80gsm8k,kim2024prometheus,zhou2024universalner} emerges as a particularly promising and practical approach, enabling much smaller models to approximate the capabilities of LLMs across a broad range of applications.

Recent work \cite{zhang2025targeteddistillationsentimentanalysis} explores target distillation for sentiment analysis. Their method employs sentiment-related \textit{instructions} and \textit{user texts} to prompt the teacher model, generating a corpus enriched with sentiment knowledge, which is then used to optimize the student model. We argue that the effectiveness of this process critically hinges on the comprehensiveness of instructions and the quantity of user texts. However, these requirements introduce two major challenges: (1) crafting sufficiently comprehensive instructions is labor-intensive, as it requires covering various perspectives for analyzing subjective content, such as polarity, emotion, lexicon, and rhetorical devices; (2) utilizing large-scale user texts incurs substantial computational costs throughout the distillation pipeline, which constrains the practical applicability of distillation-based methods.

\begin{figure}[t]
\centering
\includegraphics[width=.96\linewidth]{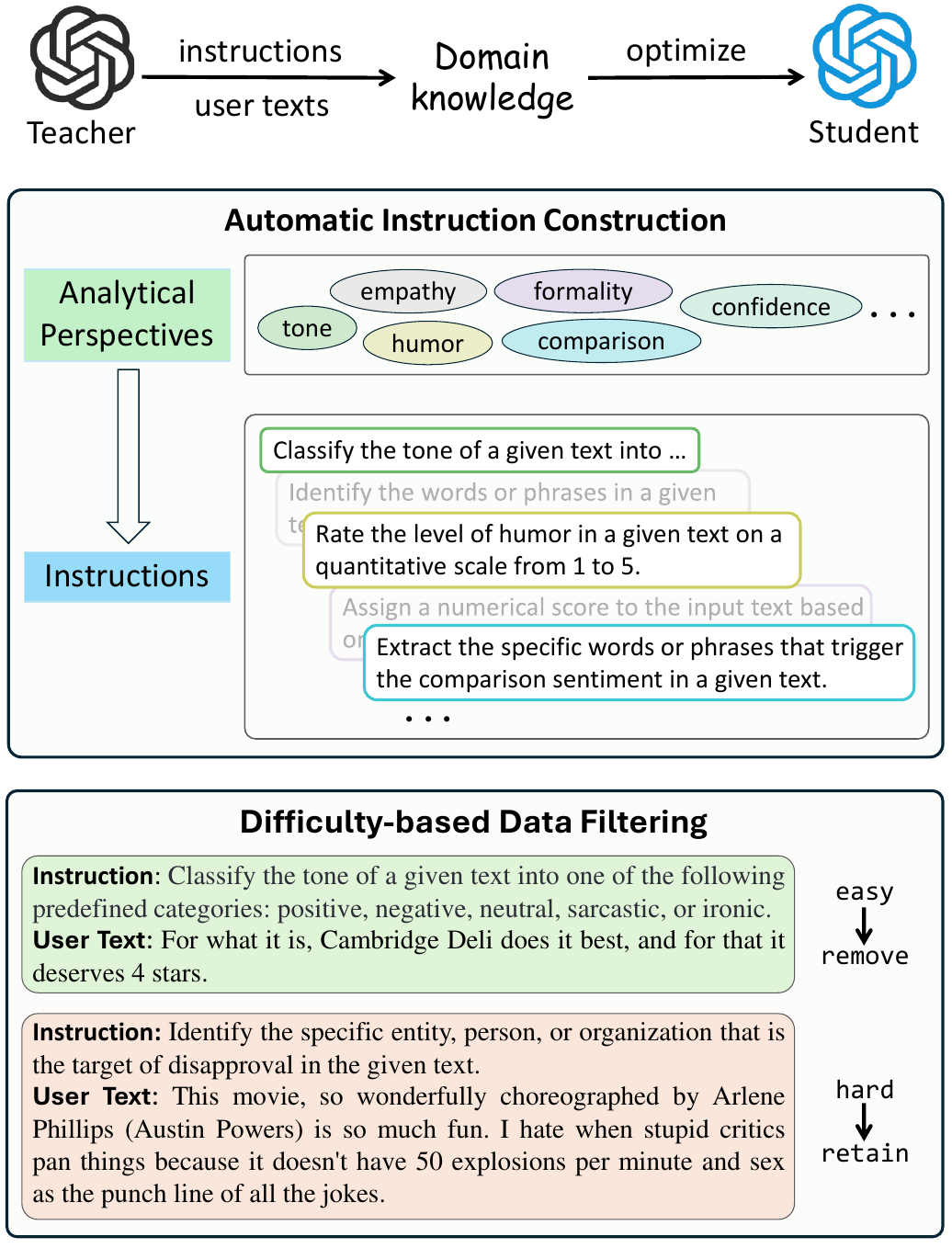}
\caption{
Illustration of our approach: (1) we extract sentiment knowledge from the teacher model through instructions and user texts and then utilize it to optimize the student model; (2) we generate diverse instructions based on various analytical perspectives to ensure comprehensive distillation; (3) we assess the difficulty of instructions and user texts and then reduce the proportion of simple samples to ensure efficient distillation.
}
\label{fig:intro}
\end{figure}

To address these two challenges, this paper introduces a comprehensive and efficient distillation framework (\textsc{CompEffDist}) for sentiment analysis. As illustrated in Figure~\ref{fig:intro}, the framework comprises two key modules. The first is attribute-based automatic instruction construction. It identifies and enumerates a wide range of sentiment-related attributes from user texts, applies clustering techniques to group these attributes into distinct analytical perspectives, and subsequently generates diverse instructions based on these analytical perspectives. As such, the module constructs comprehensive instructions without requiring labor-intensive efforts.

The second module, termed difficulty-based data filtering, aims to filter out overly simple data to boost data efficiency. This module is motivated by the hypothesis that simple data contributes minimally to model optimization \cite{liu2024what}. Specifically, we devise a ranking-based metric that employs the student model to assess the difficulty of instructions and user texts. These difficulty scores are then generalized into a proxy model, enabling efficient scoring. Finally, we apply a difficulty-prioritized sampling strategy to decrease the proportion of simple data, thereby reducing the computational cost of the distillation pipeline.

We conduct experiments across multiple model series (Llama-3, Qwen-3, and Gemma-3) and evaluate their sentiment analysis capabilities using \textsc{SentiBench} \cite{zhang2025targeteddistillationsentimentanalysis}. The experimental results reveal that:
(1) Our approach enables 3B student models to achieve performance on par with teacher models on most tasks, despite being up to 20x smaller in size. (2) Compared to baseline methods, our approach attains the same level of performance using only 10\% of the distillation data. These results highlight the effectiveness and promising potential of our approach.

\section{Preliminaries: Targeted Distillation}
\label{sec:targeted-distillation}

Targeted distillation aims to transfer domain knowledge from a teacher model $\cal T$ to a student model $\cal S$. The process generally consists of two stages. The first stage is to extract domain knowledge from the teacher model. Existing methods \cite{xu-etal-2023-inheritsumm,zhang-etal-2024-elad,kim2024prometheus,zhou2024universalner} typically utilize a large collection of instruction-user text pairs $(ins,x)$ to prompt the teacher model, generating responses $\hat{y}$:
\begin{align}
\hat{y} \sim \mathcal{T}(y\ |\ {ins}, x).
\end{align}
The resulting triples $(ins,x,\hat{y})$ are considered to encode rich domain knowledge.  In the second stage, the student model is fine-tuned on these triples using the language modeling objective, formulated as:
\begin{align}
    \max \sum_{ins,x,\hat{y}}  \mathcal{S}(\hat{y}\ |\ ins, x). 
\end{align}

\vspace{5pt}
\noindent
\textbf{Challenges.}
The effectiveness of the aforementioned process critically relies on the quality of the instructions and user texts employed. To ensure comprehensive coverage of distilled knowledge, the instruction set must capture a wide range of relevant perspectives. In sentiment analysis, these perspectives refer to various dimensions for analyzing subjective content. They may include sentiment polarity, emotion types, linguistic expressions, as well as higher-level aspects like rhetorical devices and contextual background. However, manually summarizing all such perspectives and crafting the corresponding instructions is extremely challenging and often impractical.

Simultaneously, the user text collection must span a broad spectrum of contextual variations. For example, in terms of linguistic expressions, it should include diverse patterns such as explicit sentiment words, factual statements, comparisons, metaphors, and sarcastic utterances. Achieving such diversity typically depends on the size and richness of the user text corpus, consequently increasing the computational demands of teacher model prompting and student model optimization.

\section{Comprehensive \& Efficient Distillation}

We introduce a comprehensive and efficient distillation framework (\textsc{CompEffDist}) to address the challenges of previous methods. The framework consists of two modules: (i) attribute-based automatic instruction construction, which generates diverse instructions to ensure comprehensive distillation; and (ii) difficulty-based data filtering, aiming to reduce the proportion of simple data during distillation, thereby enhancing data efficiency.

\subsection{Attribute-based Automatic Instruction Construction}

\begin{figure}[t]
\centering
\includegraphics[width=.82\linewidth]{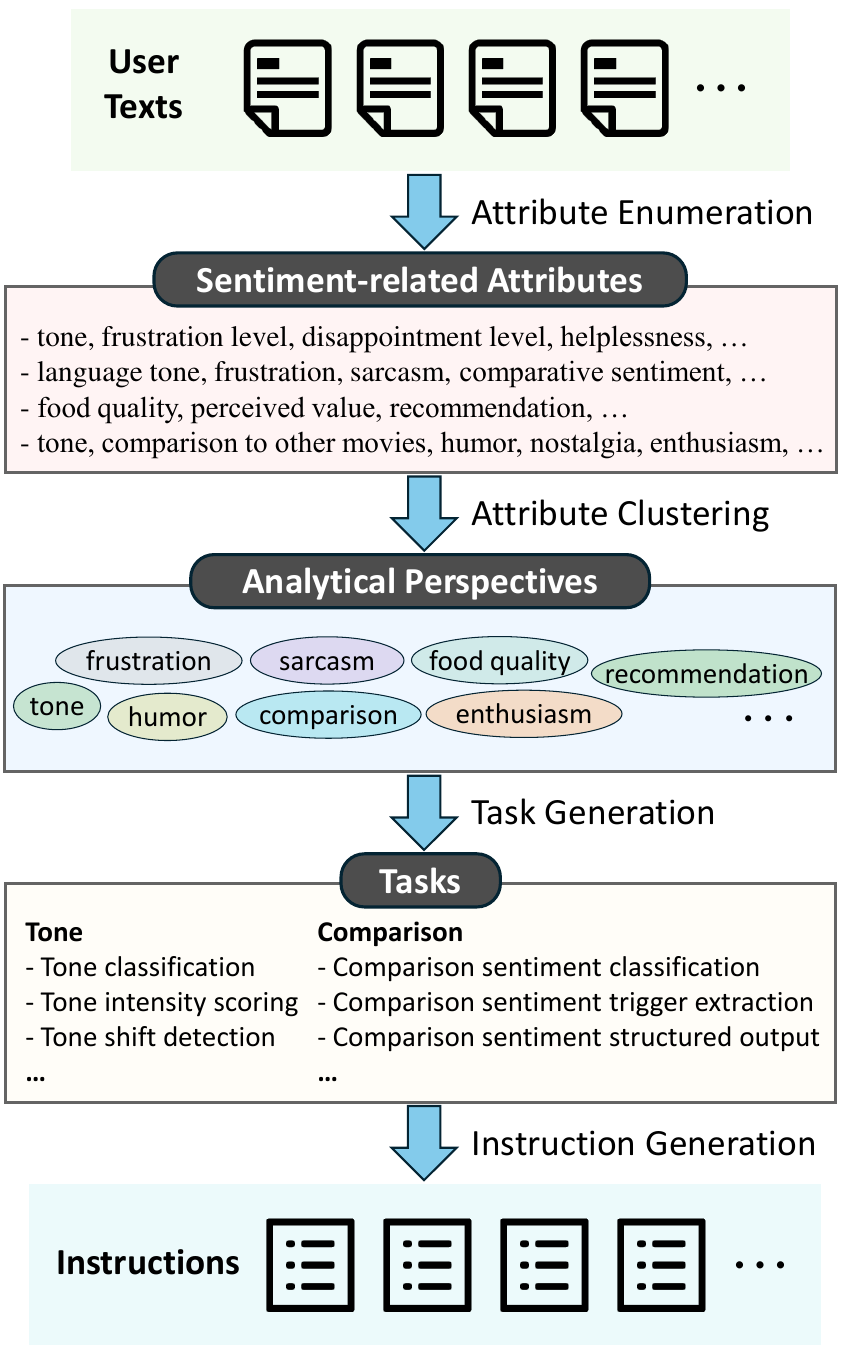}
\caption{
Illustration of attribute-based automatic instruction construction.
}
\label{fig:instruction-generation}
\end{figure}

As discussed in Section \ref{sec:targeted-distillation}, 
it is essential for instructions to cover a diverse range of analytical perspectives. Inspired by \citet{lou2024muffin}, we adopt an approach in which the teacher model identifies sentiment-related attributes from user texts, which are then used to generate specific instructions. As illustrated in Figure \ref{fig:instruction-generation}, the process comprises four steps: attribute enumeration, attribute clustering, task generation, and instruction generation.

\vspace{5pt}
\noindent
\textbf{Attribute Enumeration.}
We observe that real-world user texts inherently contain a sufficiently diverse range of sentiment-related attributes. Consider the following example:
\begin{quote}
``I wish I could give it zero stars. Whoever thinks this smells like a lemon, needs help. This is the most disgusting, repulsive, overwhelming, stinky cleaner I ever had the displeasure of using.''
\end{quote}
This text exhibits a high degree of emotional intensity and conveys strong negative sentiments such as frustration and disgust. Besides, it employs sarcasm\footnote{The rationale of sarcasm is that ``need help'' is not a genuine suggestion to seek medical attention, but rather a sarcastic critique of the other person's absurd judgment.} as a rhetorical device within the context of product dissatisfaction. In summary, this example involves the following attributes: emotional intensity, frustration, disgust, sarcasm, and product dissatisfaction.

Building on the above observation, we prompt the teacher model to identify and enumerate sentiment-related attributes present in user texts. A total of 20K user texts are used in this step. After normalizing the collected attributes, we obtain approximately 1,800 distinct attributes. The complete prompt and implementation details are provided in Appendix \ref{app:a1}.

\vspace{5pt}
\noindent
\textbf{Attribute Clustering.}
Many attributes are semantically equivalent, but differ only in form or phrasing. For example, terms such as `tone', `language tone', `tone of language', and `tone of voice' all convey the same underlying meaning.
We therefore employ clustering techniques to consolidate semantically similar attributes.
Specifically, we first employ \texttt{UAE-Large-V1}\footnote{Available at \url{https://huggingface.co/WhereIsAI/UAE-Large-V1}.} \cite{li-li-2024-aoe}, an embedding model, to project the textual attributes into vector representations. We then apply the affinity propagation algorithm \cite{doi:10.1126/science.1136800} to cluster these representations.

We ultimately obtain 180 clusters. Moving forward, we refer to these clusters as analytical perspectives and use the most frequently occurring attribute within each cluster as its name. The representative analytical perspectives include tone, comparison, humor, and food quality, covering expression styles, linguistic phenomena, and concrete aspects. A detailed presentation and analysis can be found in
Section~\ref{sec:data-analyses}.

\vspace{5pt}
\noindent
\textbf{Task \& Instruction Generation.}
For each analytical perspective, we prompt the teacher model to brainstorm a series of tasks, where each task consists of a task name and a brief description. To guide the task generation process, we provide the teacher model with predefined task types, including classification, regression, extraction, structured output, and open-ended generation. For example, under the analytical perspective of `tone', the generated tasks comprise: (i) tone classification, (ii) tone intensity scoring, (iii) tone shift detection, (iv) tone comparison to neutral, (v) tone-related entity extraction, (vi) tone profiling, and (vii) tone-based summarization.

Subsequently, we instruct the teacher model to synthesize complete instructions by enriching the task description, incorporating specific requirements, and generating a set of demonstrations. During the demonstration generation process, we make a special effort to balance the class distribution. Detailed implementation is described in Appendix \ref{app:a2}. Figure \ref{fig:instruction-example} presents an example of the resulting instruction.

\begin{figure}[h]
\centering
\includegraphics[width=1.\linewidth]{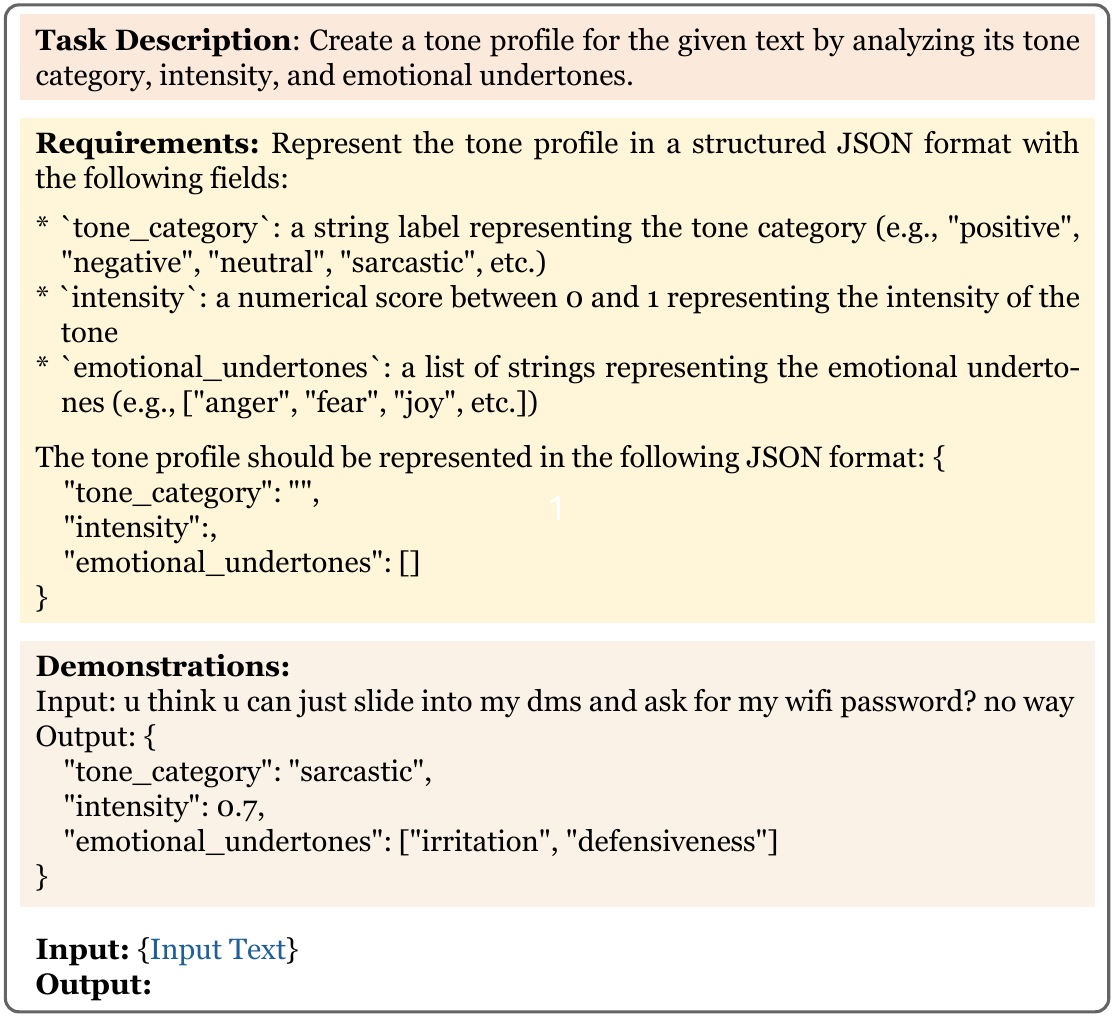}
\caption{
Generated instruction for tone profiling.
}
\label{fig:instruction-example}
\end{figure}

After obtaining a large and diverse set of instructions, we randomly pair them with user texts to construct a comprehensive collection of instruction-user text pairs. These pairs are then used to prompt the teacher model for response generation. An alternative approach is to match instructions with user texts based on their associated attributes. In our experiments, we find that both methods achieve comparable performance. Detailed results are presented in Appendix~\ref{app:d}.

\subsection{Difficulty-based Data Filtering}

As mentioned in Section~\ref{sec:targeted-distillation}, targeted distillation requires a large amount of user texts to ensure effectiveness, increasing the computational cost. To address this challenge, we assess the difficulty of instruction-user text pairs and adopt a difficulty-prioritized sampling strategy to reduce the proportion of simple data. In terms of difficulty assessment, we first evaluate each pair through a ranking-based metric. The resulting difficulty scores are optionally used to train a proxy model that enables more efficient difficulty estimation.

\vspace{5pt}
\noindent
\textbf{Ranking-based Difficulty Metric.} 
We evaluate the difficulty of a triplet $(ins,x,y)$ by assessing how well the student model $\cal S$ can reproduce the response $y$ given the input $(ins,x)$.
Most existing methods \cite{xie-etal-2024-efficient,li-etal-2024-quantity} rely on perplexity-based metrics for this purpose:
\begin{align}
{\rm PPL} = -\frac{1}{|y|}\sum_{t=1}^{|y|}\log P_{\cal S}(y_t\ |\ {ins}, x, y_{<t}).
\end{align}
However, perplexity is unsuitable for evaluating sentiment analysis samples. This is because sentiment analysis tasks typically involve categorical outputs rather than free-form text generation, and full-vocabulary probability distributions do not provide reliable indicators of correctness. Therefore, we devise a ranking-based scoring metric.

We first curate a small subset of the distillation data to warm up the student model. Subsequently, this warmed-up model is used to assess the difficulty of a given triplet $(ins, x, y)$. For each token $y_t$, we estimate the size of the relevant label space using top-$p$ sampling, denoted as $N_t$. We then determine the ranking position of $y_t$ within the top-$p$ distribution, denoted as $r_t$. Based on these values, we define the difficulty score for $y_t$ as follows:
\begin{align}
d(y_t) = 
\begin{cases}
    \dfrac{r_t - 1}{N_t} & \text{if } r_t \leq N_t, \\
    1 & \text{otherwise}.
\end{cases}
\end{align}
The overall difficulty of the response $y$ is computed as the average of the token-level difficulty scores. However, to avoid bias from format tokens (e.g., punctuation marks such as [ or "), we exclude tokens whose scores fall below a threshold $\varepsilon_d$ during the averaging process. Detailed implementations are presented in Appendix \ref{app:a3.1}.

\vspace{5pt}
\noindent
\textbf{Proxy Model.} 
The aforementioned difficulty metric requires access to the teacher response $ y $, implying that this method can only reduce the optimization cost of the student model while leaving the prompting cost of the teacher model unaffected. We therefore explore training a proxy model that does not require teacher responses, thereby allowing for data filtering before the teacher model prompting. The proxy model $\cal P$ is an autoregressive model with a regression head. It takes an instruction $ins$ and a user text $x$ as input and outputs a difficulty score $\hat{d}$:
\begin{align}
    \hat{d} &= {\cal P}(ins, x).
\end{align}
We optimize the proxy model using the mean-squared error (MSE) loss, formulated as:
\begin{align}
\mathcal{L} &= {\rm MSE}(\hat{d}, d),
\end{align}
where $d$ denotes the score derived from the ranking-based metric.

While using the proxy model can reduce both the teacher model's prompting cost and the student model's optimization cost, its performance is inferior due to the absence of teacher responses. Detailed experimental results are presented in Section \ref{sec:analysis-of-data-filtering}. Consequently, the proxy model is treated as an optional component within the data filtering module.

\vspace{5pt}
\noindent
\textbf{Difficulty-Prioritized Sampling Strategy} prioritizes more challenging samples while reducing the proportion of easily-learned samples. For a group of $M$ samples sharing the same instruction, we estimate their difficulty scores using the ranking-based metric or the proxy model. These samples are then sorted in ascending order of their scores, and each is assigned a sampling probability of $\frac{\rho - 0.5}{M}$, where $\rho$ denotes its rank. We perform stochastic sampling based on these probabilities. As a result, 50\% of the samples are expected to be retained. We also explore two variants: (i) global sampling, which ranks samples across all instructions based on their difficulty scores and adopts a unified sampling procedure; and (ii) hard-only sampling, which retains only the most difficult 50\% of samples for each instruction. Both variants yield suboptimal performance. Detailed results and analysis are provided in Section~\ref{sec:analysis-of-data-filtering}.

\begin{figure}[t]
\centering
\includegraphics[width=.84\linewidth]{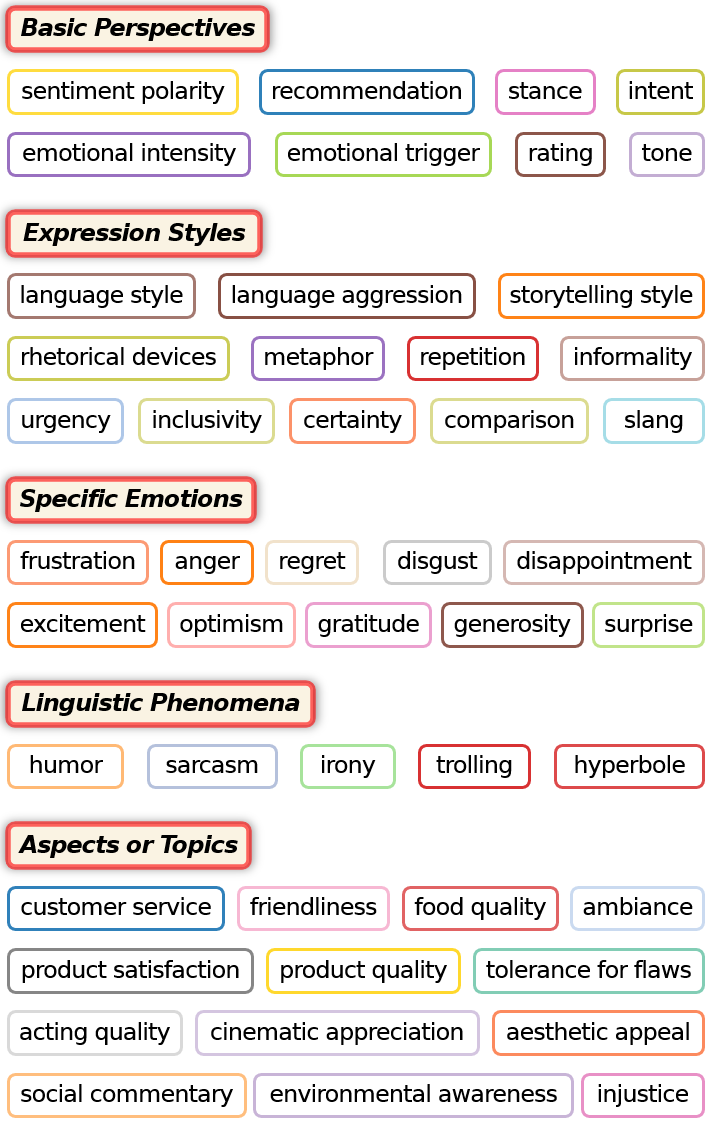}
\caption{
Visualization of the representative analytical perspectives.
A more complete visualization is provided in Figure \ref{fig:t-SNE} of Appendix \ref{app:b}.
}
\label{fig:tiny-cluster}
\end{figure}

\section{Data Analysis}
\label{sec:data-analyses}

\textbf{Visualization of Analytical Perspectives.}
The generated analytical perspectives span a broad semantic range. As shown in Figure \ref{fig:tiny-cluster}, they can be broadly categorized into five main categories: (i) basic analytical perspectives, such as sentiment polarity, emotional intensity, and tone; (ii) expression styles, such as language style, and rhetorical devices; (iii) specific emotions, such as frustration, anger, excitement, and optimism; (iv) linguistic phenomena, such as humor, sarcasm, and trolling; and (v) concrete aspects or topics, such as customer service, food quality, acting quality, and social commentary. Moreover, a certain degree of overlap or nesting among analytical perspectives can also be observed. For example, `product quality' is a sub-aspect of `product satisfaction'. We believe that this level of redundancy is potentially beneficial, as it enhances the comprehensiveness of the perspective set and supports the generation of more diverse tasks.

\begin{figure}[h]
\centering
\includegraphics[width=.9\linewidth]{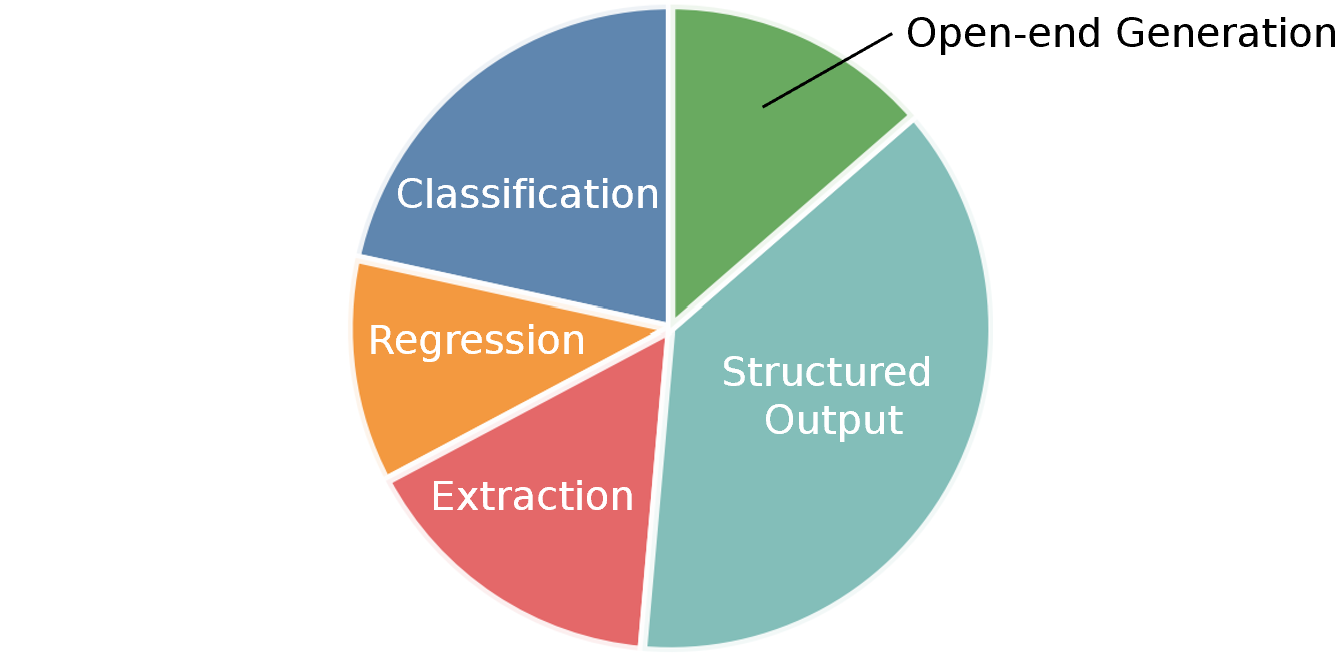}
\caption{
Type distribution of generated tasks.
}
\label{fig:task-prop}
\end{figure}

\vspace{5pt}
\noindent
\textbf{Data Statistic.}
We obtain a total of 3,707 tasks and 50K samples. Figure \ref{fig:task-prop} presents the distribution of task types. There are five task categories, with structured-output tasks comprising a notably high proportion. This is because LLMs tend to generate complex and composite tasks. We also offer the length distribution of instructions, user texts, and responses in Figure \ref{fig:task-stat} of Appendix \ref{app:b}.

\newcommand{\sixf}[1]{\fontsize{7pt}{\baselineskip}\selectfont(+#1)}
\newcommand{\sixd}[1]{\fontsize{7pt}{\baselineskip}\selectfont(-#1)}

\begin{table*}[ht]
\centering
\fontsize{8.5pt}{0.84\baselineskip}\selectfont
\setlength\tabcolsep{0.55pt}
\begin{tabular}{
p{94pt}
>{\centering\arraybackslash}m{25pt}
>{\centering\arraybackslash}m{25pt}
>{\centering\arraybackslash}m{25pt}
>{\centering\arraybackslash}m{25pt}
>{\centering\arraybackslash}m{25pt}
>{\centering\arraybackslash}m{25pt}
>{\centering\arraybackslash}m{25pt}
>{\centering\arraybackslash}m{25pt}
>{\centering\arraybackslash}m{25pt}
>{\centering\arraybackslash}m{25pt}
>{\centering\arraybackslash}m{25pt}
>{\centering\arraybackslash}m{25pt}
>{\centering\arraybackslash}m{45pt}} 
\toprule
\multirow{2}*{\textbf{Models}} & \multicolumn{4}{c}{\textbf{BSA}} & \multicolumn{4}{c}{\textbf{MSA}} & \multicolumn{4}{c}{\textbf{FSA}} & \multirow{2}*{\textbf{Avg}}\\
\cmidrule(lr){2-5}  \cmidrule(lr){6-9} \cmidrule(lr){10-13} 
& {IMDb} & {Yelp2} & {SST2} 
& {Twitter} & {Irony} & {Emoti.}  
& {Stance} & {Intim.} & {ATSA} & {ACSA} & {ASQP} & SSA \\
\midrule
EmoLlama-chat-7B  & 91.27 & 97.10 & 93.46 & 64.84 & 70.28 & 70.48 & 74.52 & 41.77 & 40.19 & 50.31 & 20.20 & 23.18 & 61.47  \\
EmoLlama-chat-13B & 93.63 & 97.87 & 94.31 & 59.87 & 65.94 & 70.75 & 73.65 & 45.28 & 48.31 & 58.92 & 23.07 & 30.83 & 63.53\\
EmoLlama-3-3B & 90.86 & 96.50 & 91.96 & 66.25 & 71.51 & 74.88 & 74.33 & 48.55 & 47.09 & 48.72 & 16.55 & 22.84 & 62.50\\
GPT-3.5 & 93.70 & 98.30 & 96.31 & 60.15 & 78.64 & 75.61 & 79.99 & 52.63 & 56.43 & 66.67 & 30.30 & 44.01 & 69.40 \\
GPT-4o & 93.91 & 98.18 & 97.13 & 70.84 & 77.58 & 76.66 & 85.22 & 53.29 & 56.72 & 72.64 & 34.75 & 51.46 & 72.37\\
\midrule
Llama-3-70B & 95.30 & 98.10 & 97.14 & 68.75 & 83.99 & 75.87 & 85.21 & 53.68 & 59.48 & 74.13 & 32.14 & 50.11 & 72.83\\
\hdashline[2pt/4pt]
Llama-3-8B  & 94.17	& 98.07 & 95.90 & 66.58 & 82.63 & 73.00 & 75.86 & 49.85 & 59.00 & 65.53 & 23.49 & 34.90 & 68.25\\
~+ \textbf{\textsc{CompEffDist} (Ours)} & 93.56 & 98.07 & 96.23 & 68.81 & 85.89 & 74.56 & 82.35 & 53.36 & 63.01 & 70.08 & 30.37 & 45.57 & 71.82\sixf{3.57} \\
\hdashline[2pt/4pt]
Llama-3-3B  & 92.57 & 96.53 & 93.59 & 61.45 & 64.00 & 68.88 & 71.43 & 33.32 & 52.74 & 53.23 & 14.33 & 23.56 & 60.47\\
~+ Distill. \textit{w/} Alpaca-data & 92.37 & 97.37 & 93.92 & 57.70 & 66.59 & 64.47 & 72.05 & 28.70 & 50.96 & 54.11 & 21.16 & 28.58 & 60.66\sixf{0.19}\\
~+ Distill. \textit{w/} Lamini-data & 92.80 & 97.33 & 94.91 & 62.07 & 70.10 & 65.61 & 72.49 & 40.28 & 53.73 & 56.46 & 19.99 & 27.25 & 62.59\sixf{2.12}\\
~+ \textsc{Know \& ICLDist} & 94.30	& 98.17	& 95.41 & 69.57 & 85.25 & 77.47 & 75.10 & 48.24 & {53.07} & {65.22} & {24.61} & {36.17} & {68.55\sixf{8.08}} \\
~+ \textbf{\textsc{CompEffDist} (Ours)} & 93.67 & 97.12 & 94.78 & 68.17 & 82.86 & 76.29 & 78.77 & 54.32 & 58.17 & 67.22 & 31.34 & 35.98 & 69.89\sixf{9.42} \\
\midrule
Qwen-3-32B & 94.37 & 97.87 & 94.86 & 62.80 & 83.06 & 73.12 & 79.28 & 52.87 & 60.36 & 72.94 & 33.80 & 49.51 & 71.24 \\
\hdashline[2pt/4pt]
Qwen-3-4B & 91.87 & 97.70 & 94.45 & 69.46 & 79.51 & 68.27 & 73.09 & 48.17 & 57.43 & 67.55 & 28.16 & 42.95 & 68.22 \\
~+ Distill. \textit{w/} Alpaca-data & 92.90 & 97.93 & 94.76 & 68.20 & 82.15 & 69.81 & 74.10 & 45.63 & 57.30 & 66.92 & 29.08 & 44.22 & 68.58\sixf{0.36}\\
~+ Distill. \textit{w/} Lamini-data & 92.43 & 98.17 & 94.16 & 68.50 & 84.02 & 64.78 & 72.89 & 50.43 & 56.11 & 66.10 & 28.93 & 38.21 & 67.89\sixd{0.33}\\
~+ \textsc{Know \& ICLDist} & 93.40 & 98.03 & 95.82 & 69.07 & 82.23 & 75.53 & 76.87 & 50.49 & 59.21 & 71.31 & 30.44 & 37.81 & 70.02\sixf{1.80} \\
~+ \textbf{\textsc{CompEffDist} (Ours)} & 92.57 & 97.83 & 94.67 & 67.98 & 84.86 & 73.09 & 77.08 & 54.03 & 60.05 & 71.47 & 32.24 & 42.46 & 70.69\sixf{2.47} \\
\midrule
Gemma-3-27B & 93.57 & 98.27 & 96.80 & 68.68 & 82.70 & 75.59 & 83.25 & 61.28 & 62.89 & 73.72 & 33.00 & 54.00 & 73.64\\
\hdashline[2pt/4pt]
Gemma-3-4B & 92.10 & 97.00 & 93.81 & 62.20 & 63.00 & 73.30 & 75.68 & 51.37 & 55.01 & 61.01 & 23.74 & 44.97 & 66.10\\
~+ Distill. \textit{w/} Alpaca-data & 92.50 & 97.40 & 94.09 & 60.09 & 75.83 & 74.46 & 74.99 & 44.60 & 54.34 & 64.08 & 24.94 & 41.98 & 66.61\sixf{0.51} \\
~+ Distill. \textit{w/} Lamini-data & 93.40 & 97.73 & 94.91 & 63.58 & 80.37 & 71.93 & 75.22 & 49.99 & 55.03 & 63.43 & 20.27 & 38.04 & 66.99\sixf{0.89} \\
~+ \textsc{Know \& ICLDist} & 93.10 & 97.87 & 95.04 & 68.12 & 72.52 & 75.41 & 76.83 & 53.82 & 58.02 & 69.82 & 32.91 & 38.93 & 69.37\sixf{3.27}\\
~+ \textbf{\textsc{CompEffDist} (Ours)} & 92.19 & 97.33 & 94.45 & 64.34 & 78.56 & 75.30 & 77.72 & 54.49 & 57.45 & 66.60 & 26.53 & 48.59 & 69.46\sixf{3.36} \\
\bottomrule
\end{tabular}
\caption{
Comparison results on \textsc{SentiBench} ($F_1$-score, \%). BSA, MSA, and FSA denote basic sentiment analysis, multi-faceted sentiment analysis, and fine-grained sentiment analysis, respectively.
\textsc{Know \& ICLDist} is trained using 300K samples, while our method uses 50K samples. EmoLlama-3-3B refers to a Llama-3-3B model fine-tuned on the same instruction dataset as EmoLlama model.
}
\label{tab:main-results}
\end{table*}

\section{Experiments}
\subsection{Experimental Setup}

\textbf{Implementation Details.}  
We conduct extensive experiments across multiple LLM series, namely Llama-3 \cite{grattafiori2024llama3herdmodels}, Qwen-3 \cite{yang2025qwen3technicalreport}, and Gemma-3 \cite{gemmateam2025gemma3technicalreport}. The specific teacher-student setups are ({Llama-3.1-70B-instruct}, Llama-3.2-3B-instruct), (Qwen-3-32B, Qwen-3-4B), and (Gemma-3-27B-it, Gemma-3-4B-it).
If not specified, analytical experiments are conducted on the Llama-3 series models.

The user text corpus for the distillation process is collected from IMDb, Yelp, Amazon, and Twitter. We use 20K user texts to generate instructions, resulting in 3,707 distinct instructions. An additional 100K user texts are then paired with these instructions to form 100K instruction-user text pairs. These pairs will be filtered based on difficulty, retaining only 50\% of them. The remaining pairs are used for distillation. Hyperparameter settings are detailed in Appendix \ref{app:a4}. After distillation, we evaluate the student models on \textsc{SentiBench} \cite{zhang2025targeteddistillationsentimentanalysis}, with dataset statistics presented in Appendix \ref{app:c}.

\vspace{5pt}
\noindent
\textbf{Baselines.}
We select three categories of baselines for comparison. The first is two generic distillation methods that fine-tune the student model using Alpaca-data \cite{alpaca} and Lamini-data \cite{wu-etal-2024-lamini}.
The second is \textsc{Know \& ICLDist} \cite{zhang2025targeteddistillationsentimentanalysis}, a two-stage approach leveraging manually written instructions to distill knowledge for sentiment analysis. The third is EmoLlama \cite{10.1145/3637528.3671552}, which fine-tunes Llama models using a multi-task affective analysis instruction dataset. In addition, we provide results of GPT-3.5 and GPT-4o\footnote{Available at \url{https://chat.openai.com/}. The specific models used are \texttt{gpt-3.5-turbo-0125} and \texttt{gpt-4o- 2024-11-20}.} as reference.

\subsection{Main Results}

Table \ref{tab:main-results} presents the comparison results between our approach and the baseline methods. These results suggest that our approach enables the student model to attain performance close to that of the teacher model on most tasks and consistently outperforms all baseline methods, demonstrating its effectiveness.

Furthermore, we make the following observations. Firstly, the performance gains from the two generic distillation methods are minimal, indicating their inefficiency in transferring specific capabilities. Secondly, our approach surpasses the previous method relying on manually constructed instructions (\textit{i.e.}, \textsc{Know \& ICLDIST}). This improvement demonstrates that our approach reduces human effort while achieving superior performance. Thirdly, there is a notable performance gap between the student and teacher models on the FSA tasks, and our approach successfully narrows this gap. We attribute this improvement to the diverse structured output tasks in the automatically generated instruction set. Finally, the performance gains on Qwen-3 and Gemma-3 are relatively modest compared to those observed on Llama-3. This can be attributed to their extensive use of knowledge distillation during training process \cite{yang2025qwen3technicalreport,gemmateam2025gemma3technicalreport}. As these models already incorporate sophisticated distillation techniques in their training pipelines, the incremental benefits of our method are naturally reduced.

\subsection{Effect of Instruction Comprehensiveness}

\begin{figure}[t]
\centering
\includegraphics[width=.9\linewidth]{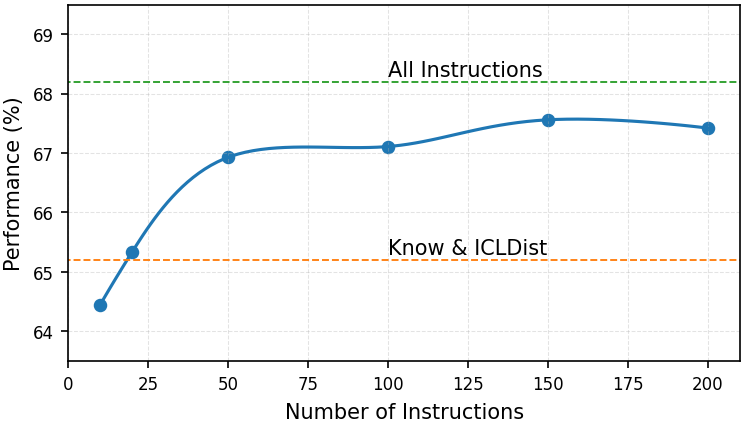}
\caption{
Performance trend of the student model with varying numbers of instructions (\%). The distillation dataset size is 20K, and data filtering is not applied.
}
\label{fig:comprehensiveness}
\end{figure}

One of the key claims of this work is that the diversity of instruction sets is a critical factor in ensuring effective distillation. We conduct exploratory experiments to assess the impact of instruction variety. Results in Figure \ref{fig:comprehensiveness} show that the student model performance gradually improves as the variety of instructions increases, providing strong support for our hypothesis. Besides, we compare our instruction set with previous work under different data budgets. As presented in Table \ref{tab:comprehensiveness}, our approach achieves results comparable to the prior method using only 20K samples---less than 10\% of 300K samples used in the prior method. This substantial reduction in data requirements highlights that the comprehensiveness of the instruction set also enhances the overall efficiency of the distillation process.

\begin{table}[ht]
\centering
\fontsize{8.5pt}{0.83\baselineskip}\selectfont
\begin{tabular}{lcc} 
\toprule
 Models & Avg-F1 & $\Delta$ \\
\midrule
Llama-3-3B & 60.47 & -\\
~+~\textsc{Know \& ICLDist} (300K) & 68.55 & +8.08 \\
~+~\textsc{Ours} (20K)  & 68.19 & +7.72 \\
~+~\textsc{Ours} (100K) & 70.17 & +9.70 \\
\bottomrule
\end{tabular}
\caption{
Performance comparison between our instruction set and that of the previous method (\%).}
\label{tab:comprehensiveness}
\end{table}

\begin{figure}[htb]
\centering
\includegraphics[width=.99\linewidth]{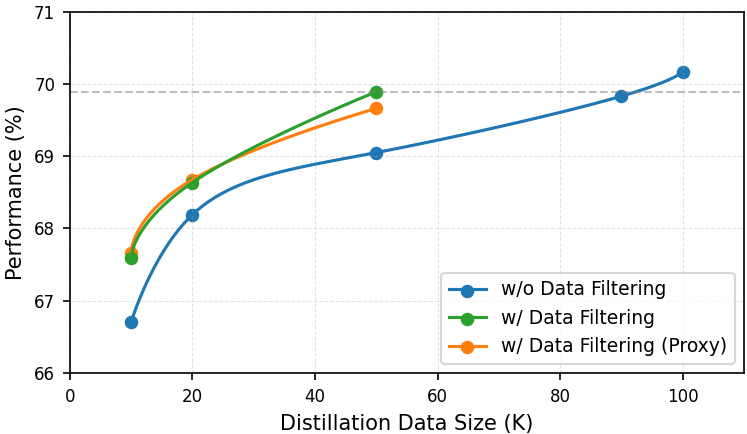}
\caption{
Performance trend of the student model with increasing data quantity (\%).
}
\label{fig:data-quantity}
\end{figure}

\subsection{Analysis of Data Filtering}
\label{sec:analysis-of-data-filtering}

Figure \ref{fig:data-quantity} illustrates the impact of data quantity on student model performance. As expected, performance improves steadily with increasing data size, highlighting the importance of sufficient distillation data. Moreover, our data filtering methods substantially enhance data efficiency, such that using only 50K filtered data can match the performance of 90K original data. This result demonstrates the effectiveness of the proposed filtering methods. Furthermore, the proxy model performs worse than direct scoring. However, it reduces both the prompting cost of the teacher model and the optimization cost of the student model. 

\begin{table}[h]
\centering
\fontsize{8.5pt}{0.83\baselineskip}\selectfont
\setlength\tabcolsep{4pt}
\begin{tabular}{lcc} 
\toprule
 Filtering Methods & Avg-F1 & $\Delta$ \\
\midrule

\textsc{Ours} & 69.89 & - \\
\hdashline[2pt/4pt]
\multicolumn{1}{l}{\textit{Difficulty Metric Ablations}} \\
Perplexity  & {69.09} & -0.80 \\
IFD  & {69.08} & -0.81\\
User Text Length  & {69.21} & -0.68\\
\hdashline[2pt/4pt]
\multicolumn{1}{l}{\textit{Sampling Strategy Ablations}} \\
Global Sampling   &  69.29 & -0.60\\
Hard-only Sampling & 69.23 & -0.69\\
\bottomrule
\end{tabular}
\caption{
Ablation studies of data filtering methods (\%). For difficulty metric ablations, we use difficulty-prioritized sampling strategy. For sampling strategy ablations, we use the ranking-based difficulty metric.
}
\label{tab:filtering}
\end{table}

We conduct ablation experiments on the filtering methods. Firstly, we compare different metrics for difficulty assessment: perplexity, IFD \cite{li-etal-2024-quantity}, and text length. Results in Table \ref{tab:filtering} show that these metrics perform worse than our ranking-based metric. Secondly, we explore variants of our sampling strategy. We find that both global sampling and hard-only sampling result in suboptimal performance. We attribute the poor performance of global sampling to the large variation in instruction difficulty, which can lead to over-selection of certain instruction types and thus reduce diversity. As for hard-only sampling, we believe that restricting the distillation data to only difficult samples hinders the learning of the student model.

\subsection{Further Analysis}
We have the following further analyses in the Appendix:
\begin{itemize}
    \setlength{\itemsep}{1pt}
    \item \hyperref[para:further1]{Analysis of Instruction-User Text Pairing.}
    \item \hyperref[para:further2]{Data Filtering on Other Baselines.}
    \item \hyperref[para:further3]{Results on Complex Contexts.}
    \item \hyperref[para:further4]{Case Study of Difficulty Assessment.}
\end{itemize}

\section{Related Work}
\textbf{Targeted Distillation}.
Knowledge distillation techniques have been widely applied to develop more accessible and compact models \cite{alpaca, vicuna2023, wu-etal-2024-lamini}. Targeted distillation, which focuses on transferring LLMs' capabilities in specific applications, has recently gained significant attention. Existing methods can be broadly categorized into two paradigms. The first \cite{ding-etal-2023-gpt, he-etal-2024-annollm, xu-etal-2023-inheritsumm, zhou2024universalner} treats the LLM as an annotator, generating large-scale task-specific pseudo-labels for training a smaller model. This method typically employs limited instructions and is effective mainly for narrowly defined tasks. The second \cite{zhang2025targeteddistillationsentimentanalysis, kim2024prometheus} constructs a broader set of instructions to transfer LLMs' capabilities across a targeted domain. While offering stronger effectiveness and broader generalization, it also imposes higher demands on the quantity and diversity of instructions.

\vspace{5pt}
\noindent
\textbf{Instruction Generation} has emerged as a key research direction, due to its critical role in improving the coverage of distilled knowledge. Existing methods can be broadly categorized into two types.
The first \cite{wang2023selfinstructaligninglanguagemodels, honovich-etal-2023-unnatural, xu2024wizardlm} adopts a bootstrap strategy, generating new instructions based on existing ones. However, this method requires a large seed instruction set and often suffers from limited diversity. The second is attribute-based methods \cite{wu-etal-2024-lamini, lou2024muffin}, generating instructions by specifying topics, entities, or text segments. Its main challenge lies in developing a high-quality and diverse attribute set.
To address this, we identify a large number of attributes from user texts and employ clustering algorithms to group them into meaningful analytical perspectives.

\vspace{5pt}
\noindent
\textbf{Data Selection} has been extensively studied, especially as model sizes continue to grow, leading to prohibitively high fine-tuning and inference costs. The main criteria guiding data selection include diversity, quality, and difficulty. A few studies explore manual curation of instruction data \cite{köpf2023openassistantconversationsdemocratizing, zhou2023lima}, but such methods are labor-intensive and less scalable. More recent efforts have therefore focused on automatic selection methods.
For diversity, techniques such as vocabulary coverage, semantic tagging, and clustering are employed \cite{cao2024instruction, lu2024instag, ge-etal-2024-clustering}. For quality, filtering based on advanced LLMs is a common practice \cite{chen2024alpagasus, slimorca}. For difficulty, most existing methods rely on the student model's uncertainty, with ongoing efforts aimed at developing more robust and reliable difficulty metrics \cite{li-etal-2024-quantity, kung2023active}. In this paper, we highlight that current difficulty metrics are not well-suited for sentiment analysis tasks. To address this, we propose a ranking-based metric.

\section{Conclusions}

To develop lightweight sentiment analysis models, we introduce \textsc{CompEffDist}, a comprehensive and efficient distillation framework. This framework automatically generates a large and diverse set of instructions via an attribute-based method and applies difficulty-based data filtering to boost data efficiency. Leveraging this framework, we construct a dataset containing 3,707 distinct tasks and 50K samples. Applying it to knowledge distillation, we enable 3B student models to achieve performance comparable to that of 20x larger teacher models on most tasks. Furthermore, our approach attains results on par with baseline methods using only 10\% of the data, demonstrating its superior data efficiency.

\section*{Limitations}
We discuss potential limitations of this work:
\begin{itemize} 
\item 
\textsc{CompEffDist} does not include task-level deduplication or filtering operations. The large number of generated tasks inevitably contains overlaps and some low-quality instances. Introducing task-level deduplication and quality-based filtering could increase the proportion of high-quality, long-tail tasks, thereby improving data efficiency in the distillation process. However, identifying task overlaps and assessing instruction quality remain challenging.
\item 
\textsc{CompEffDist} does not incorporate quality control for the teacher model's responses. Teacher models can generate incorrect or biased outputs, which can be transferred to the student model and affect its performance. Incorporating quality assurance techniques, such as reflection, reasoning, or consistency checks, has the potential to improve the effectiveness and reliability of knowledge distillation. However, this would also introduce additional computational costs. Balancing the trade-off between cost and performance improvement is an important direction for future research.
\end{itemize} 
We believe that these limitations point to promising directions for future research.

\section*{Ethics Statement}
Large language models for sentiment analysis have enabled progress in areas such as public health and commercial applications; yet their reliance on large-scale pretraining corpora raises ethical concerns, including risks of privacy violations, cultural and annotator subjectivity, and systematic harms to marginalized groups \cite{ehics-sheet-for-sa}. While knowledge distillation substantially improves efficiency and deployability, prior work shows that it can also transfer and intensify existing biases, exacerbating disparities across sentiment classes and demographic subgroups. 

Accordingly, ethical evaluation of distilled sentiment models should not only emphasize improvements in overall performance but also recognize the risks of propagating biases and exacerbating disparities across categories and social subgroups \cite{distillationfairness}. Therefore, the community should place greater emphasis on assessing subgroup- and category-level fairness, accompanied by clearer documentation of risks and limitations. In addition, exploring fairness-aware distillation methods and developing practical guidelines could help mitigate potential misuse in sensitive or high-stakes applications.

\section*{Acknowledgments}
This work was supported by the National Natural Science Foundation of China 62176076 and 62576120,  Natural Science Foundation of Guang Dong 2023A1515012922, the Major Key Project of PCL2023A09,  CIPSC-SMP-ZHIPU Large Model Cross-Disciplinary Fund ZPCG20241119405 and Key Laboratory of Computing Power Network and Information Security, and Ministry of Education under Grant No.2024ZD020.

\bibliographystyle{acl_natbib}

\appendix

\newpage
\section*{Organization of Appendices}
We organize the appendix into four sections:
\begin{itemize} 
\item Appendix \ref{app:a} presents additional implementation details of our method;
\item Appendix \ref{app:b} provides more comprehensive data visualization;
\item Appendix \ref{app:c} describes the evaluation setup and dataset statistics;
\item Appendix \ref{app:d} offers further analysis of our method.
\end{itemize}

\section{Further Implementation Details}
\label{app:a}

\subsection{Attribute Enumeration and Clustering}
\label{app:a1}

We leverage the teacher model to identify and enumerate sentiment-relevant attributes from user texts. The complete prompt used for this step is shown in Table~\ref{tab:attr-prompt}. By parsing the model responses, we obtain a large number of attributes, which are then standardized to construct an attribute pool. Attributes that appear fewer than or equal to 10 times are removed, resulting in a total of 1,785 distinct attributes.

These attributes are subsequently mapped into a vector space using UAE embeddings. We apply affinity propagation clustering to group the vectors. The hyperparameters are set as follows: percentile\_rate = 0.5 and damping = 0.9. To incorporate attribute frequency into the clustering process, we first map the frequency counts $x$ using the following transformation:
\begin{align}
    y = 1 + \log (1 + x),
\end{align}
and then replicate each attribute $ y $ times before performing clustering.

\begin{table*}[htbp]
\centering
\fontsize{8.5pt}{0.82\baselineskip}\selectfont
\setlength\tabcolsep{1pt}
\begin{tabular}{p{15.5cm}} 
\toprule

Instruction: Given the following input, what kind of sentiment-related attributes does it have? \\
\\
Requirements: \\
\\
1. Please brainstorm as many related attributes as possible.\\
2. Be creative. Any interesting perspectives are welcome!\\
3. Each attribute should typically reflect a particular characteristic of the input text.\\
4. Each attribute should be followed with Attribute Description (a brief description of what the attribute represents) and Attribute Value (the corresponding attribute value as reflected in the text).\\
5. Feel free to ignore the tedious and specific content. Just focus on some general textual attributes!\\
\\
Input: {\color{blue}\{Input Text\}}\\
\\
Attribute:\\
\bottomrule
\end{tabular}
\caption{
The prompt for attribute enumeration.
}
\label{tab:attr-prompt}
\end{table*}

\begin{table*}[htbp]
\centering
\fontsize{8.5pt}{0.82\baselineskip}\selectfont
\setlength\tabcolsep{1pt}
\begin{tabular}{p{15.5cm}} 
\toprule

\textbf{Open-end Generation Task Generation}\\

Please generate prompts for analyzing subjective texts such as product reviews or social media according to the following rules: \\
1. Each prompt should capture the core and commonalities of the following attribute categories and without relying on specific attribute: {\color{blue}\{Perspective\}}. \\
\hspace*{1em}    - The explanation for {\color{blue}\{Attribution1\}} is {\color{blue}\{Brief Explaination of Attribution1\}}.\\
\hspace*{1em}    - The explanation for {\color{blue}\{Attribution2\}} is {\color{blue}\{Brief Explaination of Attribution2\}}.\\
\hspace*{1em}    - The explanation for {\color{blue}\{Attribution3\}} is {\color{blue}\{Brief Explaination of Attribution3\}}.\\
\hspace*{1em}    - The explanation for {\color{blue}\{Attribution4\}} is {\color{blue}\{Brief Explaination of Attribution4\}}.\\
\hspace*{1em}    - The explanation for {\color{blue}\{Attribution5\}} is {\color{blue}\{Brief Explaination of Attribution5\}}.\\
2. Ensure that each prompt is domain-general by using neutral references such as "this text" avoiding any specific domain indications.\\
3. Each prompt should be designed to help better understand subjective texts by deconstructing it based on the specified attribute categories.\\
4. Employ diverse strategies, which may include but are not limited to:\\
\hspace*{1em}    - Open-ended deconstruction instructions\\
\hspace*{1em}    - Diagnostic questions\\                         
5. Ensure that your responses are structured in ordered numbers.\\
\\
Generated prompt: \\
\midrule

\textbf{Constrained Task Generation}\\

I want you to focus on the following text attribute: **{\{\color{blue}Perspective\}}({\{\color{blue}Brief Explaination of Perspective\}})**, and systematically generate a diverse range of tasks that target a single text. Please make sure each task includes the following elements:\\
\hspace*{1em}    - Task Name: a concise title that captures the core goal or theme of the task.\\
\hspace*{1em}    - Task Description: an explanation of the problem this task aims to solve or the objective it aims to achieve.\\
The task types should be diverse, such as: \\
1. Classification\\
\hspace*{1em}    - Closed-set categories classification\\
\hspace*{1em}    - Open-ended categories classification\\
2. Scoring or Rating\\
\hspace*{1em}    - Quantitative scales\\
3. Information Extraction\\
\hspace*{1em}    - Keywords, key sentences, triggers\\
\hspace*{1em}    - Root causes, contextual dependencies, and more\\
4. Structured Output\\
\hspace*{1em}    - JSON, tables, or other machine-readable formats\\
\hspace*{1em}    - Potentially includes multiple fields (roles, attribute values, etc.)\\
When designing these tasks, please follow these guidelines:\\
\hspace*{1em}    - Clarity: Each task's goal should be described methodically.\\
\hspace*{1em}    - Diversity: Aim for a wide range of creative ideas across classification, scoring, extraction, and extended analyses.\\
\hspace*{1em}    - All tasks must target a single text. Therefore, do not generate tasks involving comparisons between two texts.\\
Based on the above requirements, please list several diverse tasks focused on **\{Attribution\}**.\\
Present your output in the following structured JSON format, ensuring that it can be directly parsed.\\
\bottomrule
\end{tabular}
\caption{
The prompts for task generation.
}
\label{tab:task-generation}
\end{table*}

\begin{table*}[htbp]
\centering
\fontsize{8.5pt}{0.82\baselineskip}\selectfont
\setlength\tabcolsep{1pt}
\begin{tabular}{p{15.5cm}} 
\toprule
\textbf{Instruction Generation}\\
Please rewrite the task based on the task name and description, making the task definition more standardized and normalized.\\
\\
Task Name: {\color{blue}\{Task Name\}}\\
Task Description:{\color{blue}\{Task Description\}}\\
\\
Below are the specific requirements and guidelines: \\
1. Avoiding Ambiguity: Ensure task description, requirement and constraint is precise, complete, and free of ambiguity. If the task contains two direction, specify one direction in the task description and requirments and you should NOT add any requirments in input. \\
\\
2. Ensure the rewritten task is consistent with the original task description.\\
\\
3. Task Elements: Ensure that each task definition includes the following components:\\
\hspace*{1em}    - Task Name: A concise title of the task.\\
\hspace*{1em}    - Task Description: A detailed explanation of the task and should contain the following parts: \\
\hspace*{2em}        - Explicitly specifying the expected output format and requirements (e.g., classification label, numerical score, structured JSON, Python list).\\
\hspace*{2em}        - If the task is a classification task or contains classification task as subtask, for closed-set classification, you should explicitly list all allowed labels. For open-set classification, you should instruct the model to infer the appropriate labels from the input.\\
\hspace*{2em}        - If the task is a annotation/extraction task, you should specify whether the extracted or annotated text must exactly match the original text or if modifications are allowed.\\
\hspace*{2em}        - If the task requires structured output, specify the exact structure (for example, a JSON schema or Python list format) and enumerate all required fields.\\
\hspace*{1em}    - Task Examples: You should provide at least EIGHT concrete examples, each including:\\
\hspace*{2em}        - Task Input: Formatted according to the input specifications.\\
\hspace*{2em}        - Task Output: Formatted according to the output specifications.\\
\bottomrule
\end{tabular}
\caption{
The prompts for instruction generation.
}
\label{tab:instruction-generation}
\end{table*}

\begin{table*}[htbp]
\centering
\fontsize{8.5pt}{0.82\baselineskip}\selectfont
\setlength\tabcolsep{1pt}
\begin{tabular}{p{15.5cm}} 
\toprule
\textbf{Demo Generation}\\

Generate two instances for the following task. The text part in the samples needs to refer to the style, vocabulary, and themes in the Reference Texts. Carefully read the task description to ensure the correct labeling in the generated samples.\\
\\
Reference Texts:\\
{\color{blue}\{Reference Text1\}} \\
{\color{blue}\{Reference Text2\}} \\
\\
Task Description:\\
{\color{blue}\{Task Description\}} \\
\\
Give your response in the following format: \\
Input: \{\} \\
Output: \{\} \\

\bottomrule
\end{tabular}
\caption{
The prompts for demo generation.
}
\label{tab:demo-generation}
\end{table*}

\subsection{Task and Instruction Generation}
\label{app:a2}

For each analytical perspective, we prompt the teacher model to generate two types of tasks: open-ended generation tasks and constrained tasks. The corresponding prompts are provided in Table~\ref{tab:task-generation}.

For each constrained task, we further guide the model to synthesize complete instructions by enriching the descriptions and adding specific requirements. The detailed prompt for this step is shown in Table \ref{tab:instruction-generation}. In addition, we generate 32 demonstrations for each task, using the prompting templates listed in Table \ref{tab:demo-generation}.
During demonstration generation, we provide reference texts to enhance diversity. After generation, we analyze the distribution of demonstration categories. If the distribution is imbalanced, we generate additional examples for underrepresented categories to ensure a more balanced composition.

\subsection{Difficulty Assessment}
\label{app:a3}

\subsubsection{Detailed Calculation of Difficulty Metric}
\label{app:a3.1}

We compute the difficulty of a sample using the ranking-based metric. To adapt the student model to the data distribution, we first perform a warm-up phase using 5,000 distillation samples. For each token in the response, we estimate the size of the label space using top-$p$ sampling, with $p$ empirically set to 0.95. When aggregating the scores across tokens, we exclude those tokens whose scores are below a threshold $\varepsilon_d = \num{1e-6}$. However, to avoid division by zero, we ensure that at least one token is retained for each sample.

The following example illustrates the detailed calculation of our ranking-based difficulty metric, including the estimation of $N_t$ and the overall calculation process. Given a triplet $(instr, x, y)$, the ranking-based difficulty score for each target token $y_t$ is calculated through the following steps:
\begin{itemize}
\setlength{\itemsep}{1pt}  
    \item \textbf{Instruction}: ``Classify the sentiment of the following review as Positive, Negative, or Neutral.''
    \item \textbf{Input ($x$)}: ``This product is a complete waste of money. I regret buying it.''
    \item \textbf{Ground-truth label ($y$)}: ``Negative''
\end{itemize}

\noindent
\textbf{Step 1: Model Output Distribution}

\noindent
The model generates the following probability distribution over candidate label tokens:

\begin{table}[h]
\centering
\begin{tabular}{|c|c|}
\hline
\textbf{Token} & \textbf{Probability} \\
\hline
Pos & 0.45 \\
Neu & 0.40 \\
Neg & 0.11 \\
Mixed & 0.02 \\
Other tokens & 0.02 \\
\hline
\end{tabular}
\caption{Model probability distribution over candidate label tokens}
\label{tab:prob_distribution}
\end{table}

\noindent
\textbf{Step 2: Top-$p$ Sampling ($p = 0.95$)}

\noindent
We calculate cumulative probabilities in descending order:
\begin{itemize}
\setlength{\itemsep}{1pt}   
    \item `Pos': 0.45
    \item `Pos' + `Neu': 0.45 + 0.40 = 0.85
    \item `Pos' + `Neu' + `Neg': 0.85 + 0.11 = 0.96
\end{itemize}

\noindent
Since the cumulative probability first exceeds 0.95 with the inclusion of `Neg', the candidate set contains three tokens: \{`Pos', `Neu', `Neg'\}, resulting in $N_t = 3$. After sorting by predicted probability, the ground-truth token `Neg' receives rank $r_t = 3$.

\vspace{5pt}
\noindent
\textbf{Step 3: Difficulty Score Calculation}

\noindent
Since the target token appears in the candidate set ($r_t \leq N_t$), we apply the first case of the formula:

\begin{equation}
d(y_t) = \frac{r_t - 1}{N_t} = \frac{3 - 1}{3} = \frac{2}{3} \approx 0.67.
\end{equation}

To illustrate the maximum difficulty scenario, consider a case where the ground-truth token `Neg' does not appear in the top-95\% probability mass. In this situation, $r_t > N_t$, and the difficulty score becomes:

\begin{equation}
d(y_t) = 1.
\end{equation}

\noindent
This maximum score indicates that the correct label is not among the model's most probable predictions, representing the highest level of prediction difficulty.

The ranking-based difficulty metric provides an intuitive measure of prediction difficulty:
\begin{itemize}
\setlength{\itemsep}{1pt}  
    \item \textbf{Lower scores} (closer to 0): The correct token has high predicted probability and low rank, indicating easier prediction.
    \item \textbf{Higher scores} (closer to 1): The correct token has low predicted probability and high rank, indicating more difficult prediction.
    \item \textbf{Maximum score} (exactly 1): The correct token is not among the top-$p$ candidates, representing maximum difficulty.
\end{itemize}

\begin{table}[h]
\centering
\fontsize{8.5pt}{0.82\baselineskip}\selectfont

\begin{tabular}{lc} 
\toprule
\textbf{Hyper-parameter} & \textbf{Value} \\ 
\midrule
Batch Size & 3 \\ 
Learning Rate & 1.5e-4 \\ 
Training Epoch & 3 \\ 
Learning Rate Deacy & Linear \\ 
Rank & 64 \\ 
Alpha & 16 \\ 
Target Module & k\_proj,q\_proj,v\_proj,o\_proj \\ 
\bottomrule
\end{tabular}
\caption{
Hyperparameters for the proxy model's optimization.
}
\label{tab:proxy-model-hp}
\end{table}

\subsubsection{Proxy Model}
The proxy model is implemented as an autoregressive model with an additional regression head. It is initialized from the student model, \textit{i.e.}, Llama-3.2-3B-instruct. We train the proxy model on a dataset of 50K samples using LoRA \cite{hu2022lora}, with hyperparameters specified in Table~\ref{tab:proxy-model-hp}.

\subsection{Knowledge Distillation}
\label{app:a4}

In the process of constructing distillation samples, each instruction is paired with multiple randomly sampled user texts. Furthermore, we randomly sample 1 to 8 demonstrations from the demonstration pool. The instruction, selected demonstrations, and user text are then fed into the teacher model to generate a response. The resulting samples are subsequently used to optimize the student model, with the maximum sequence length set to 2048. The optimization hyperparameters for the three student models are listed in Tables~\ref{tab:llama-hp}, \ref{tab:qwen-hp}, and \ref{tab:gemma-hp}, respectively.

\begin{table}[h]
\centering
\fontsize{8.5pt}{0.82\baselineskip}\selectfont

\begin{tabular}{lc} 
\toprule
\textbf{Hyper-parameter} & \textbf{Value} \\ 
\midrule
Batch Size & 128 \\ 
Learning Rate & 5e-5 \\ 
Training Epoch & 4 \\ 
Learning Rate Deacy & Cosine \\ 
Warmup Step Ratio & 0 \\ 
Weight Decay & 0.1 \\ 
Adam $\beta_1$ & 0.9 \\ 
Adam $\beta_2$ & 0.999 \\ 
\bottomrule
\end{tabular}
\caption{
Hyperparameters for Llama-3.2-3B-instruct.
}
\label{tab:llama-hp}
\end{table}

\begin{table}[h]
\centering
\fontsize{8.5pt}{0.82\baselineskip}\selectfont

\begin{tabular}{lc} 
\toprule
\textbf{Hyper-parameter} & \textbf{Value} \\ 
\midrule
Batch Size & 128 \\ 
Learning Rate & 2e-5 \\ 
Training Epoch & 4 \\ 
Learning Rate Deacy & Cosine \\ 
Warmup Step Ratio & 0.05 \\ 
Weight Decay & 0.1 \\ 
Adam $\beta_1$ & 0.9 \\ 
Adam $\beta_2$ & 0.999 \\ 
\bottomrule
\end{tabular}
\caption{
Hyperparameters for Qwen-3-4B.
}
\label{tab:qwen-hp}
\end{table}

\begin{table}[h]
\centering
\fontsize{8.5pt}{0.82\baselineskip}\selectfont

\begin{tabular}{lc} 
\toprule
\textbf{Hyper-parameter} & \textbf{Value} \\ 
\midrule
Batch Size & 128 \\ 
Learning Rate & 1e-5 \\ 
Training Epoch & 4 \\ 
Learning Rate Deacy & Cosine \\ 
Warmup Step Ratio & 0.05 \\ 
Weight Decay & 0.01 \\ 
Adam $\beta_1$ & 0.9 \\ 
Adam $\beta_2$ & 0.999 \\ 
\bottomrule
\end{tabular}
\caption{
Hyperparameters for Gemma-3-4B-it.
}
\label{tab:gemma-hp}
\end{table}

\section{Further Data Analysis}
\label{app:b}

\begin{figure}[t]
\centering
\includegraphics[width=.64\linewidth]{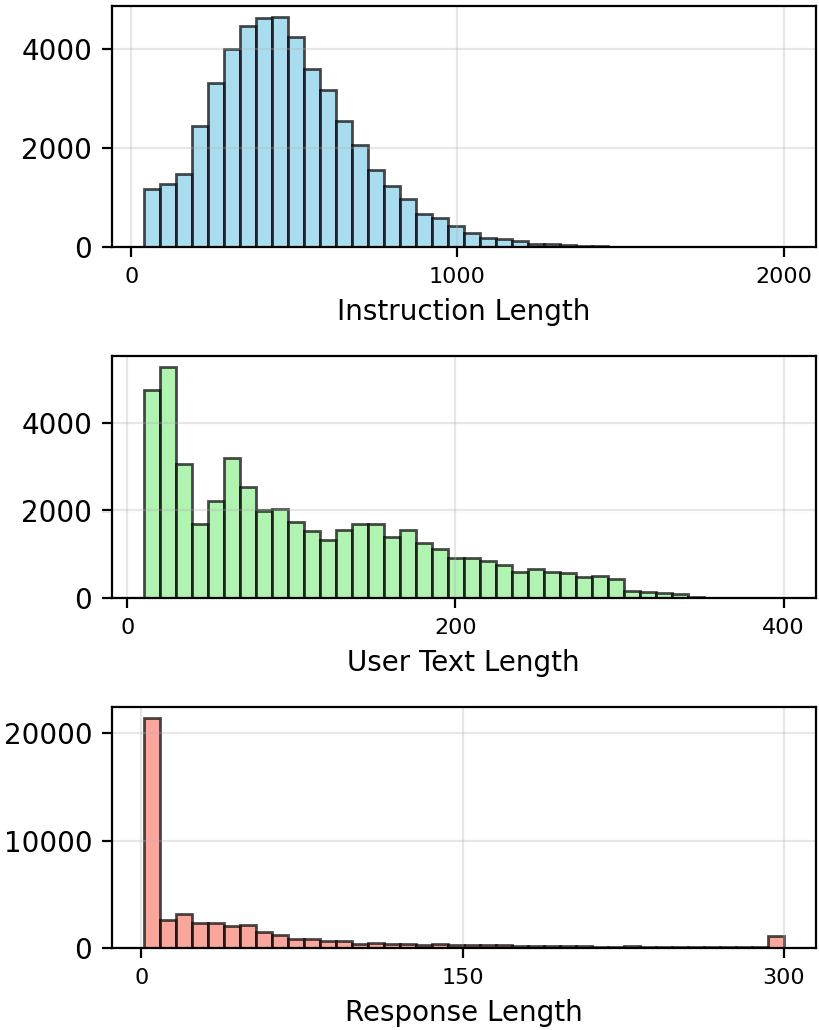}
\caption{
Length distribution in the distillation dataset.
}
\label{fig:task-stat}
\end{figure}

\begin{figure*}[t]
\centering
\includegraphics[width=.99\linewidth]{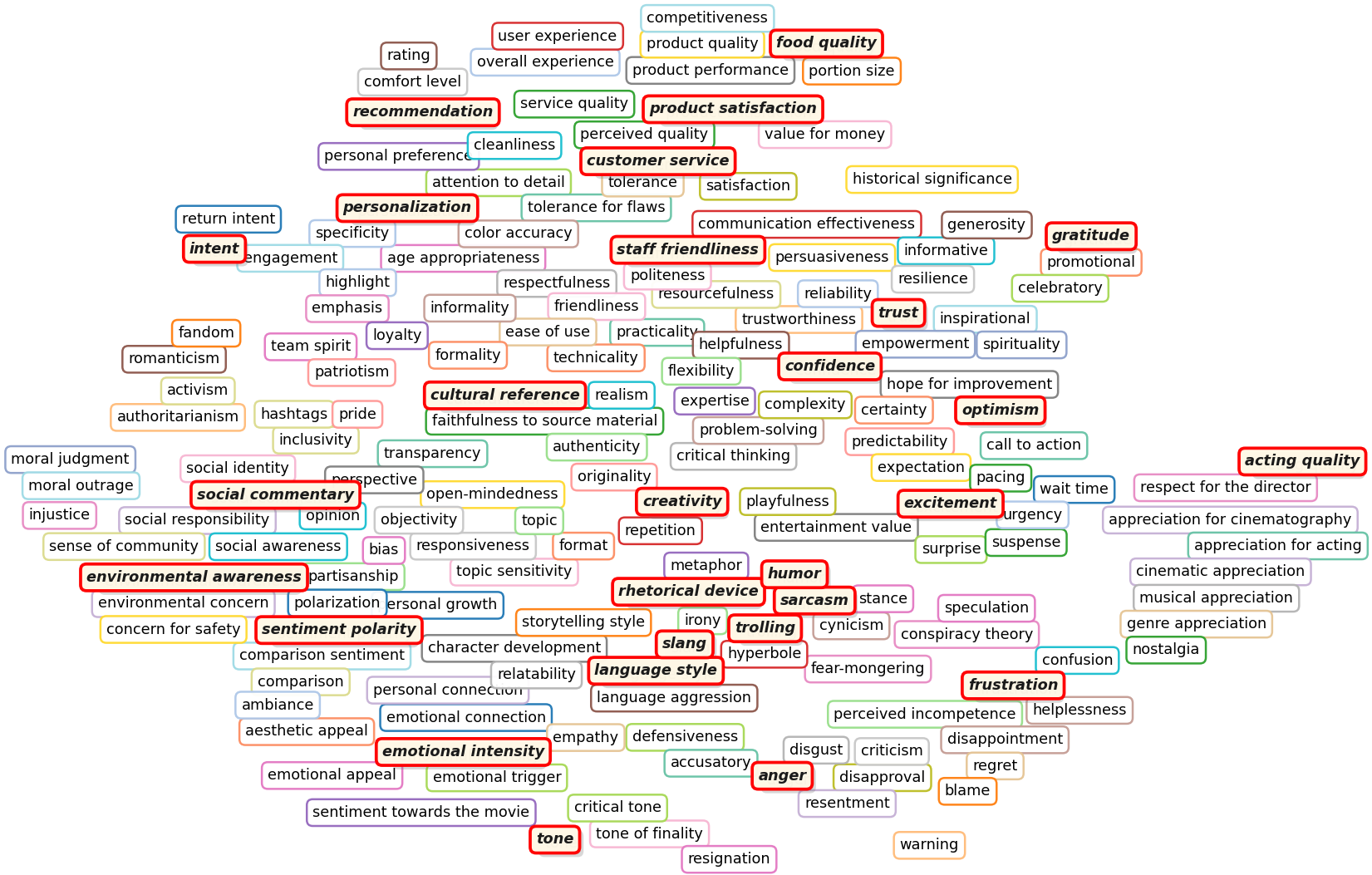}
\caption{
A t-SNE \cite{JMLR:v9:vandermaaten08a} visualization of the generated analytical perspectives using the \texttt{UAE} embeddings \cite{li-li-2024-aoe}. Representative perspectives are highlighted with red bounding boxes. For clarity, overly long names have been appropriately shortened (e.g., \textit{sense of helplessness} $\rightarrow$ \textit{helplessness}). 
}
\label{fig:t-SNE}
\end{figure*}

We visualize all the obtained perspectives in Figure~\ref{fig:t-SNE}.
Besides, we provide length statistics for the 50K samples in Figure \ref{fig:task-stat}.

\definecolor{lightgray}{gray}{0.95}
\begin{table*}[ht]
\centering
\fontsize{8.5pt}{0.82\baselineskip}\selectfont
\begin{tabular}{l c ccc c ccc c ccc c ccc c c} 
\toprule
\textbf{Task} & \textbf{Dataset} & \textbf{Train} & \textbf{Dev} & \textbf{Test} & \textbf{\#Class}\\
\midrule 
\noalign{\smallskip}
\multicolumn{7}{c}{\textsc{Basic Sentiment Analysis}} \\
\hdashline[2pt/4pt]
& IMDb  & 3000 & 300 & 1000 & 2 \\
        \multirow{-2}{*}{{Document-level sentiment classification}}
& Yelp2 & 3000 & 300 & 1000 & 2 \\
    \rowcolor{lightgray}
& SST2  & 3000 & 300 & 1821 & 2 \\
    \rowcolor{lightgray}
        \multirow{-2}{*}{{Sentence-level sentiment classification}}
& Twitter17 & 3000 & 300 & 1000 & 3 \\
\midrule
    \noalign{\smallskip}
\multicolumn{7}{c}{\textsc{Multifaceted Sentiment Analysis}} \\
\hdashline[2pt/4pt]
Irony detection & Irony18 & 3000 & 300 & 784 & 2\\
    \rowcolor{lightgray}
Emotion recognition & Emotion20 & 3000 & 300 & 1421 & 4 \\
Stance detection & P-Stance & 3000 & 300 & 2157 & 3  \\
    \rowcolor{lightgray}
Intimacy analysis & \textsc{Mint}-English & 1287 & 300 & 396 & 3  \\
\midrule
    \noalign{\smallskip}
\multicolumn{7}{c}{\textsc{Fine-Grained Sentiment Analysis}} \\
\hdashline[2pt/4pt]
Aspect term sentiment analysis & Rest16 & 1600 & 400 & 676 & -  \\
    \rowcolor{lightgray}
Aspect category sentiment analysis & Rest16 & 1600 & 400 & 676 & - \\
Aspect sentiment quad prediction
& Rest16 & 1264 & 316 & 544 & -\\
    \rowcolor{lightgray}
Structured sentiment analysis
& Opener & 1744 & 249 & 499 & - \\
\bottomrule
\end{tabular}
\caption{
dataset statistics of \textsc{SentiBench}.
}
\label{tab:sentibench}
\end{table*}

\section{Evaluation Settings}
\label{app:c}

Following the previous work \cite{zhang2025targeteddistillationsentimentanalysis}, we evaluate the models on \textsc{SentiBench} using an in-context learning setup. The dataset statistics are shown in Table~\ref{tab:sentibench}. The number of demonstrations is fixed at 4. We select demonstrations from the validation set using three different random seeds and report the average result of three runs. The prompts used are the same as those in \citet{zhang2025targeteddistillationsentimentanalysis}, except for four datasets under the FSA category. For these datasets, we refine the prompts and update the performance of the baseline models accordingly. The refined prompts are presented in Table~\ref{tab:fsa-prompt}.

\begin{table*}[t]
\centering
\fontsize{8.5pt}{0.84\baselineskip}\selectfont
\setlength\tabcolsep{0.55pt}
\begin{tabular}{
p{94pt}
>{\centering\arraybackslash}m{25pt}
>{\centering\arraybackslash}m{25pt}
>{\centering\arraybackslash}m{25pt}
>{\centering\arraybackslash}m{25pt}
>{\centering\arraybackslash}m{25pt}
>{\centering\arraybackslash}m{25pt}
>{\centering\arraybackslash}m{25pt}
>{\centering\arraybackslash}m{25pt}
>{\centering\arraybackslash}m{25pt}
>{\centering\arraybackslash}m{25pt}
>{\centering\arraybackslash}m{25pt}
>{\centering\arraybackslash}m{25pt}
>{\centering\arraybackslash}m{45pt}} 
\toprule
\multirow{2}*{\textbf{Models}} & \multicolumn{4}{c}{\textbf{BSA}} & \multicolumn{4}{c}{\textbf{MSA}} & \multicolumn{4}{c}{\textbf{FSA}} & \multirow{2}*{\textbf{Avg}}\\
\cmidrule(lr){2-5}  \cmidrule(lr){6-9} \cmidrule(lr){10-13} 
& {IMDb} & {Yelp2} & {SST2} 
& {Twitter} & {Irony} & {Emoti.}  
& {Stance} & {Intim.} & {ATSA} & {ACSA} & {ASQP} & SSA \\
\midrule

Llama-3-3B  & 92.57 & 96.53 & 93.59 & 61.45 & 64.00 & 68.88 & 71.43 & 33.32 & 52.74 & 53.23 & 14.33 & 23.56 & 60.47\\
~+ \textsc{Know \& ICLDist} & 94.30	& 98.17	& 95.41 & 69.57 & 85.25 & 77.47 & 75.10 & 48.24 & {53.07} & {65.22} & {24.61} & {36.17} & {68.55\sixf{8.08}} \\

~+ Data Filtering  & 94.70 & 98.20 & 95.66 & 69.91 & 83.93 & 77.78 & 74.76 & 45.91 & 53.10 & 65.56 & 26.90 & 32.22 & 68.22\sixf{7.75} \\

\bottomrule
\end{tabular}
\caption{
Performance comparison of the \textsc{Know\&ICLDist} baseline trained on the full 300k dataset versus 150k filtered dataset.
}
\label{tab:data-filter-on-knowicldist}
\end{table*}

\begin{table}[ht]
\centering
\fontsize{8.5pt}{0.83\baselineskip}\selectfont
\begin{tabular}{lcc} 
\toprule
 Models & Avg-F1 & $\Delta$ \\
\midrule
Llama-3-3B & 60.47 & -\\
~+~\textsc{Dist} \textit{w/} Random-Pairing  & 68.19 & +7.72 \\
~+~\textsc{Dist} \textit{w/} Attribute-Matched-Pairing  & 67.61 & +7.14 \\
\bottomrule\end{tabular}
\caption{
Comparison between two instruction-user text pairing methods (\%).}
\label{tab:pairing-comparison}
\end{table}

\section{Further Analysis}
\label{app:d}

\labelpara{para:further1}
\textbf{Analysis of Instruction-User Text Pairing.} 
We compare two strategies for pairing instructions and user texts: (i) random pairing and (ii) attribute-based matching. As shown in Table~\ref{tab:pairing-comparison}, both methods achieve similar performance, with random pairing even showing a slight advantage. We attribute this outcome to the fact that random pairing leads to a more balanced class distribution in the resulting dataset, whereas attribute-based matching tends to introduce an excessive number of positive samples. For example, in the sarcasm detection task, attribute-based matching results in an overrepresentation of sarcastic samples and an underrepresentation of non-sarcastic ones. Based on these analyses, we adopt the random pairing strategy in our final framework.

\newcommand{\fivef}[1]{\fontsize{6.5pt}{\baselineskip}\selectfont(+#1)}

\begin{table}[htb]
\centering
\fontsize{8pt}{0.83\baselineskip}\selectfont
\setlength\tabcolsep{1.6pt}
\begin{tabular}{lcccc} 
\toprule
 Models & TSA-R14 & TSA-L14 & ASA-R16 & ASA-L16 \\
\midrule
\multicolumn{5}{c}{\textsc{Implicit Sentiment Samples}}\\
\hdashline[2pt/4pt]
GPT-3.5 & 43.11 & 30.73 & 52.75 & 29.25\\
Llama-3-70B & 50.08 & 42.89 & 63.39 & 44.30\\
Llama-3-3B & 22.65 & 21.45 & 40.46 & 17.13\\
~+~\textsc{Ours} & 37.93\fivef{15.28} & 30.28\fivef{8.83} & 53.33\fivef{12.87} & 27.94\fivef{10.81} \\
\midrule
\multicolumn{5}{c}{\textsc{Multiple Sentiments Samples}}\\
\hdashline[2pt/4pt]
GPT-3.5 & 48.35 & 35.07 & 52.23 & 32.54\\
Llama-3-70B & 54.40 & 49.31 & 60.13 & 44.37\\
Llama-3-3B & 28.32 & 20.45 & 36.22 & 14.96 \\
~+~\textsc{Ours} & 43.47\fivef{23.02} & 35.73\fivef{15.28} & 51.00\fivef{14.78} & 22.30\fivef{7.34} \\
\bottomrule\end{tabular}
\caption{
Experimental results in complex contexts ($F_1$-score, \%).}
\label{tab:complex}
\end{table}

\vspace{5pt}
\labelpara{para:further2}
\noindent
\textbf{Data Filtering on Other Baselines.}
We apply our data filtering method to the \textsc{KNOW\&ICLDist} baseline to investigate its generalizability. The results in Table \ref{tab:data-filter-on-knowicldist} demonstrate the effectiveness and robustness of our method. Notably, performance degradation is minimal even when the dataset is reduced by 50\%.

\vspace{5pt}
\labelpara{para:further3}
\noindent
\textbf{Results on Complex Contexts.}
Complex contexts refer to texts that contain implicit sentiment and express multiple sentiment polarities simultaneously. We evaluate the impact of distillation on the student model's ability to perform sentiment analysis in complex contexts. The evaluation is conducted on the dataset introduced by \citet{zhang2024distillingfinegrainedsentimentunderstanding}, under an in-context learning setup with 4 demonstrations. The results in Table~\ref{tab:complex} reveal the following: (1) Llama-3-3B performs significantly worse than Llama-3-70B on both types of complex context; (2) Our approach leads to substantial improvements in the performance of the Llama-3-3B model, with average gains of 11.95\% and 15.11\% across the two settings. These findings demonstrate that our approach can effectively enhance the student model's capability to handle complex contextual understanding.

\vspace{5pt}
\labelpara{para:further4}
\noindent
\textbf{Case Study of Difficulty Assessment.} We present two representative examples of difficulty assessment in Table~\ref{tab:case}. Based on these cases, we make the following observations. Firstly, perplexity is not an effective indicator of a sample's true difficulty. As shown in the table, two samples with similar perplexity scores exhibit noticeably different levels of difficulty. Secondly, for relatively easier samples, both the ranking-based metric and the proxy model assign low difficulty scores, suggesting that their estimations are reasonably accurate in such cases. Thirdly, for more complex tasks, the proxy model tends to overestimate the difficulty. This is because the proxy model does not have access to the teacher model's response and thus cannot accurately determine whether it can replicate the teacher's output. In summary, effectively and efficiently estimating the difficulty of a sample remains a challenging problem. We believe this is a promising direction for future research.

\begin{table}[ht]
\centering
\fontsize{8.5pt}{0.82\baselineskip}\selectfont
\setlength\tabcolsep{1pt}
\begin{tabular}{p{7.5cm}} 
\toprule

\textbf{Instruction:} Classify the level of satisfaction expressed in a given text into one of the following predefined categories: Very Satisfied, Satisfied, Neutral, Dissatisfied, Very Dissatisfied. The output should be a single classification label.  \\
Expected Output Format: A single string label from the following set: ["Very Satisfied", "Satisfied", "Neutral", "Dissatisfied", "Very Dissatisfied"]\\
\textbf{Input:} The first time we watched this movie we were all sitting in a ball on the couch! Over all a very nice horror movie, if you want to get scared! We all know the scary sound of Kayako's throat sound! My son's bedroom door creaks and sounds like it and creeps him out! I think It's one of the best horror movies we own! 5 *'s! \\
\textbf{Output:} Very Satisfied\\
\textbf{Perplexity:} 1.0009 \\
\textbf{Ranking-based Metric:} 0 \\
\textbf{Proxy Model:} 0.0479 \\
\midrule
\textbf{Instruction:} Analyze the input text and provide a JSON output containing the sentiment analysis results. The output should include the following fields:...\\
\textbf{Input:} Input: Award winning bakery indeed!!! I was searching for key lime pie in the Orlando area and read multiple reviews regarding Yvette's story. Impressed by all the awards she won as a new baker lead me to give her sweets a try. HANDS DOWN, Her key lime pie is the BEST! Sweet, creamy, zest filled, homemade crust goodness will keep you coming back for more! \\
\textbf{Output:}\\
\{\\
\ \ \ \ "sentiment": "Positive",\\
\ \ \ \ "sentiment\_intensity": 5,\\
\ \ \ \ "sentiment\_triggers": [\\
\ \ \ \ \ \ \ \ "Award winning",\\
\ \ \ \ \ \ \ \ "HANDS DOWN",\\
\ \ \ \ \ \ \ \ "the BEST",\\
\ \ \ \ \ \ \ \ "Sweet, creamy, zest filled, homemade crust goodness"\\
\ \ \ \ ]\\
\}\\
\textbf{Perplexity:} 1.0814 \\
\textbf{Ranking-based Metric:} 0.1488 \\
\textbf{Proxy Model:} 0.5234 \\
\bottomrule
\end{tabular}
\caption{
Representative examples for difficulty assessment.
}
\label{tab:case}
\end{table}

\begin{table*}[htbp]
\centering
\fontsize{8.5pt}{0.82\baselineskip}\selectfont
\setlength\tabcolsep{1pt}
\begin{tabular}{p{15.5cm}} 
\toprule
\textbf{FSA - ATSA - Rest16}\\

Please perform Aspect Term Sentiment Analysis task. Given the sentence, extract all aspect terms and their corresponding sentiment polarities.\\
Return your answer in JSON format as an array of objects, each with the fields:\\
\hspace*{1em}- "aspect\_term": the extracted aspect\\
\hspace*{1em}- "sentiment": one of "positive", "negative" or "neutral\\
Example output format:

\begin{flushleft}
\textnormal{[\{"aspect\_term": "aspect\_term", "sentiment": "sentiment"\}]}
\end{flushleft} \\

\midrule
\textbf{FSA - ACSA - Rest16}\\

Please perform aspect-level sentiment analysis task. Given the sentence, tag all aspect categories and their corresponding sentiment polarities.\\
Aspect category should be selected from ["ambience general", "drinks prices", "drinks quality", "drinks style\_options", "food prices", "food quality", "food style\_options", "location general", "restaurant general", "restaurant miscellaneous", "restaurant prices", "service general"], and sentiment should be selected from ["negative", "neutral", "positive"]. \\
Return your answer in JSON format as an array of objects, each with the fields:\\
\hspace*{1em}- "aspect\_category": the selected aspect category\\
\hspace*{1em}- "sentiment": the sentiment polarity\\
If there are no aspect-sentiment pairs, return an empty list.\\
Example output format:
\begin{flushleft}
\textnormal{[\{"aspect\_category": "aspect\_category", "sentiment": "sentiment"\}]}
\end{flushleft} \\

\midrule
\textbf{FSA - ASQP - Rest16}\\

Please perform Aspect Sentiment Quad Prediction task. Given the sentence, extract all (aspect term, aspect category, opinion, sentiment polarity) quadruples.\\
1. Aspect category should be selected from ["ambience general", "drinks prices", "drinks quality", "drinks style\_options", "food general", "food prices", "food quality", "food style\_options", "location general", "restaurant general", "restaurant miscellaneous", "restaurant prices", "service general"].\\
2. Sentiment polarity should be selected from ["negative", "neutral", "positive"].\\
3. If there is no aspect term, use "NULL" as the aspect term. Only aspect term can be "NULL", aspect category, opinion and sentiment polarity CANNOT be "NULL".\\
Return your answer in JSON format as an array of objects, each with the fields:\\
\hspace*{1em}- "aspect\_term": the extracted aspect term (or "NULL")\\
\hspace*{1em}- "aspect\_category": the selected aspect category\\
\hspace*{1em}- "opinion": the expressed opinion\\
\hspace*{1em}- "sentiment": the sentiment polarity\\
Example output format:
\textnormal{[\{"aspect\_term": "aspect\_term", "sentiment": "sentiment", "opinion": "opinion", "sentiment": "sentiment"\}]}\\
\midrule
\textbf{FSA - SSA - Opener}\\

Please perform the Structured Sentiment Analysis task. Given a sentence, extract all opinion tuples in the format (holder, target, sentiment expression, sentiment polarity).\\
Each tuple should contain:\\
\hspace*{1em}- Holder: The entity expressing the sentiment, if there is no explicit holder, use "NULL" as the holder.\\
\hspace*{1em}- Target: The entity being evaluated, if there is no explicit target, use "NULL" as the target.\\
\hspace*{1em}- Sentiment Expression: The phrase conveying the sentiment, if there is no sentiment expression, use "NULL".\\
\hspace*{1em}- Sentiment Polarity: The polarity of the sentiment, either positive, negative, or neutral, if there is no sentiment expression, use "NULL".\\
Follow these rules:\\
1. If there is no sentiment expression, return "NULL" for all fields.\\
2. Return your answer in JSON format as an array of objects, each with the fields:\\
\hspace*{1em}- "holder"\\
\hspace*{1em}- "target"\\
\hspace*{1em}- "sentiment\_expression"\\
\hspace*{1em}- "sentiment\_polarity"\\
Example output format:
\textnormal{[\{"holder": "holder", "target": "target", "sentiment\_expression": "sentiment\_expression", "sentiment\_polarity": "sentiment\_polarity"\}]}\\
\bottomrule
\end{tabular}
\caption{
The refined prompts for fine-grained sentiment analysis (FSA) task.
}
\label{tab:fsa-prompt}
\end{table*}

\end{document}